\documentclass[sigconf]{acmart}% \documentclass[sigconf,screen,review,anonymous]{acmart}
\AtBeginDocument{%
  }

\setcopyright{acmlicensed}
\copyrightyear{2026}
\acmYear{2026}
\acmDOI{XXXXXXX.XXXXXXX}
\acmConference[MM '26]{the 34rd ACM International Conference on Multimedia}{November 10--14,
  2026}{Rio de Janeiro, Brazil}
\usepackage{cleveref}
\crefname{table}{Table}{Tables}
\crefname{figure}{Figure}{Figures}
\usepackage{colortbl}
\usepackage{array}
\usepackage{amsmath, bm}
\usepackage{multirow}
\usepackage{makecell}
\usepackage{graphicx}
\usepackage{algorithmicx,algorithm}
\usepackage[dvipsnames,svgnames]{xcolor}
\usepackage{xcolor}
\usepackage{svg}
\usepackage{alltt}
\usepackage{caption}
\usepackage{wrapfig}
\usepackage{pifont}       % \ding{xx}
\usepackage{bbding}       % \Checkmark,\XSolid,... (需要和pifont宏包共同使用)
\usepackage{fontawesome}  % \faCheck,\faTimes
\usepackage{tikz}
\usepackage{booktabs}
\usepackage{hyperref}

\definecolor{pink}{HTML}{DAA2B6}
\definecolor{brownnew}{HTML}{AA5757}
\definecolor{greennew}{HTML}{91CD68} 
\definecolor{bluenew}{HTML}{5EA8ED}
\definecolor{orangenew}{HTML}{E6AC83}

\usepackage[table]{xcolor} % 支持表格上色
\definecolor{lightblue}{RGB}{240, 240, 240} % 浅蓝色 (RGB)
\definecolor{pubblue}{RGB}{50, 50, 250} % 紫色
\newcommand{\tablecite}[3]{#1 {\color{pubblue}\scriptsize (#2)} \cite{#3}}
\newcommand{\B}[1]{\textbf{#1}}

\newcommand{\cmark}{\ding{51}} % 勾
\newcommand{\xmark}{\ding{55}} % 叉
\setcopyright{acmlicensed}
\copyrightyear{2026}
\acmYear{2026}
\acmDOI{XXXXXXX.XXXXXXX}% 录用后再填
\acmISBN{978-1-4503-XXXX-X/2018/06}
\begin{document}

%%
%% The "title" command has an optional parameter,
%% allowing the author to define a "short title" to be used in page headers.
\title{Frequency-Aware Semantic Fusion with Gated Injection for AI-generated Image Detection}

%%
%% The "author" command and its associated commands are used to define
%% the authors and their affiliations.
%% Of note is the shared affiliation of the first two authors, and the
%% "authornote" and "authornotemark" commands
%% used to denote shared contribution to the research.
\author{Shuchang Zhou}
\affiliation{%
  \institution{University of Electronic Science and Technology of China}
  \city{Chengdu, Sichuan}
  \country{China}}
\email{sczhou@std.uestc.edu.cn}
% \authornotemark[1]

\author{Shangkun Wu}
\affiliation{%
  \institution{University of Electronic Science and Technology of China}
  \city{Chengdu, Sichuan}
  \country{China}}
\email{2023150901020@std.uestc.edu.cn}

\author{Jiwei Wei}
\authornote{Corresponding author.}
\affiliation{%
  \institution{University of Electronic Science and Technology of China}
  \city{Chengdu, Sichuan}
  \country{China}}
\email{mathematic6@gmail.com}

\author{Ke Liu}
\affiliation{%
  \institution{University of Electronic Science and Technology of China}
  \city{Chengdu, Sichuan}
  \country{China}}
\email{liuke3068@gmail.com}

\author{Ran Ran}
\affiliation{%
  \institution{University of Electronic Science and Technology of China}
  \city{Chengdu, Sichuan}
  \country{China}}
\email{ranran@std.uestc.edu.cn}

\author{Caiyan Qin}
\affiliation{%
  \institution{Harbin Institute of Technology}
  \city{Shenzhen, Guangdong}
  \country{China}}
\email{qincaiyan@hit.edu.cn}

\author{Yang Yang}
\affiliation{%
  \institution{University of Electronic Science and Technology of China}
  \city{Chengdu, Sichuan}
  \country{China}}
\email{yang.yang@uestc.edu.cn}
%%
%% By default, the full list of authors will be used in the page
%% headers. Often, this list is too long, and will overlap
%% other information printed in the page headers. This command allows
%% the author to define a more concise list
%% of authors' names for this purpose.
% \renewcommand{\shortauthors}{Trovato et al.}

%%
%% The abstract is a short summary of the work to be presented in the
%% article.
\begin{abstract}

AI-generated images are becoming increasingly realistic and diverse, posing significant challenges for generalizable detection. While Vision Foundation Models (VFMs) provide rich semantic representations and frequency-based methods capture complementary artifact cues, existing approaches that combine these modalities still suffer from limited generalization, with notable performance degradation on unseen generative models. We attribute this limitation to two key factors: frequency shortcut bias toward easily distinguishable cues associated with specific generators and cross-domain representation conflict between high-level semantics and low-level frequency patterns. To address these issues, we propose a Frequency-aware Gated Injection Network (FGINet) to improve generalization. Specifically, we design a Band-Masked Frequency Encoder (BMFE) that applies cross-band masking in the frequency domain to reduce reliance on generator-specific patterns and encourage more diverse and generalizable representations. We further introduce a Layer-wise Gated Frequency Injection (LGFI) mechanism to progressively inject frequency cues into the VFM backbone with adaptive gating, aligning with its hierarchical abstraction and alleviating representation conflict. Moreover, we propose a Hyperspherical Compactness Learning (HCL) framework with a cosine margin objective to learn compact and well-separated representations. Extensive experiments demonstrate that FGINet achieves state-of-the-art performance and strong generalization across multiple challenging datasets.
\end{abstract}

%%
%% The code below is generated by the tool at http://dl.acm.org/ccs.cfm.
%% Please copy and paste the code instead of the example below.
%%
\begin{CCSXML}
<ccs2012>
   <concept>
       <concept_id>10010147.10010178.10010224</concept_id>
       <concept_desc>Computing methodologies~Computer vision</concept_desc>
       <concept_significance>500</concept_significance>
       </concept>
   <concept>
       <concept_id>10002978.10003029.10003032</concept_id>
       <concept_desc>Security and privacy~Social aspects of security and privacy</concept_desc>
       <concept_significance>500</concept_significance>
       </concept>
 </ccs2012>
\end{CCSXML}

\ccsdesc[500]{Computing methodologies~Computer vision}
\ccsdesc[500]{Security and privacy~Social aspects of security and privacy}

%%
%% Keywords. The author(s) should pick words that accurately describe
%% the work being presented. Separate the keywords with commas.
\keywords{AI-generated image detection, Frequency-band masking, Gated injection}
%% A "teaser" image appears between the author and affiliation
%% information and the body of the document, and typically spans the
%% page.
% \begin{teaserfigure}
%   \includegraphics[width=\textwidth]{sampleteaser}
%   \caption{Seattle Mariners at Spring Training, 2010.}
%   \Description{Enjoying the baseball game from the third-base
%   seats. Ichiro Suzuki preparing to bat.}
%   \label{fig:teaser}
% \end{teaserfigure}

% \received{20 February 2007}
% \received[revised]{12 March 2009}
% \received[accepted]{5 June 2009}

%%
%% This command processes the author and affiliation and title
%% information and builds the first part of the formatted document.
\maketitle

\section{Introduction}

The rapid evolution of AI-generated images~\cite{goodfellow2014gan,ho2020ddpm,dhariwal2021adm,rombach2022sd,Peebles2023DiT} has enabled the creation of highly realistic and diverse visual content, posing significant challenges to the reliability of digital media. In response, AI-generated image detection has attracted increasing attention. Vision Foundation Models (VFMs), such as DINOv3~\cite{siméoni2025dinov3}, have recently demonstrated strong potential due to their powerful open-set semantic representation capability. Meanwhile, prior studies~\cite{frank2020leveraging, tan2024rethinking, durall2020watchup, chu2025fire} reveal that generative processes inevitably introduce subtle yet systematic artifacts in the frequency domain, providing complementary cues for distinguishing real and synthetic images.

Existing frequency-based detection methods can be broadly categorized into three paradigms. The first directly exploits statistical patterns or high-frequency components as discriminative cues~\cite{frank2020leveraging, tan2024rethinking, li2025improving}. The second leverages frequency reconstruction discrepancies to capture deviations from natural image formation~\cite{wang2023dire,zhuang25mrcl,chu2025fire}. More recently, a third line of work integrates frequency modeling into VFMs~\cite{yansanity,cheng2025co,zhou2025aigi}, aiming to combine high-level semantic consistency with low-level artifact cues. By leveraging the strong generalization capability of VFMs and the sensitivity of frequency signals to generative artifacts, these approaches have the potential to better distinguish visually realistic yet statistically inconsistent images.

\begin{figure*}[t]
  \centering
\includegraphics[width=\textwidth]{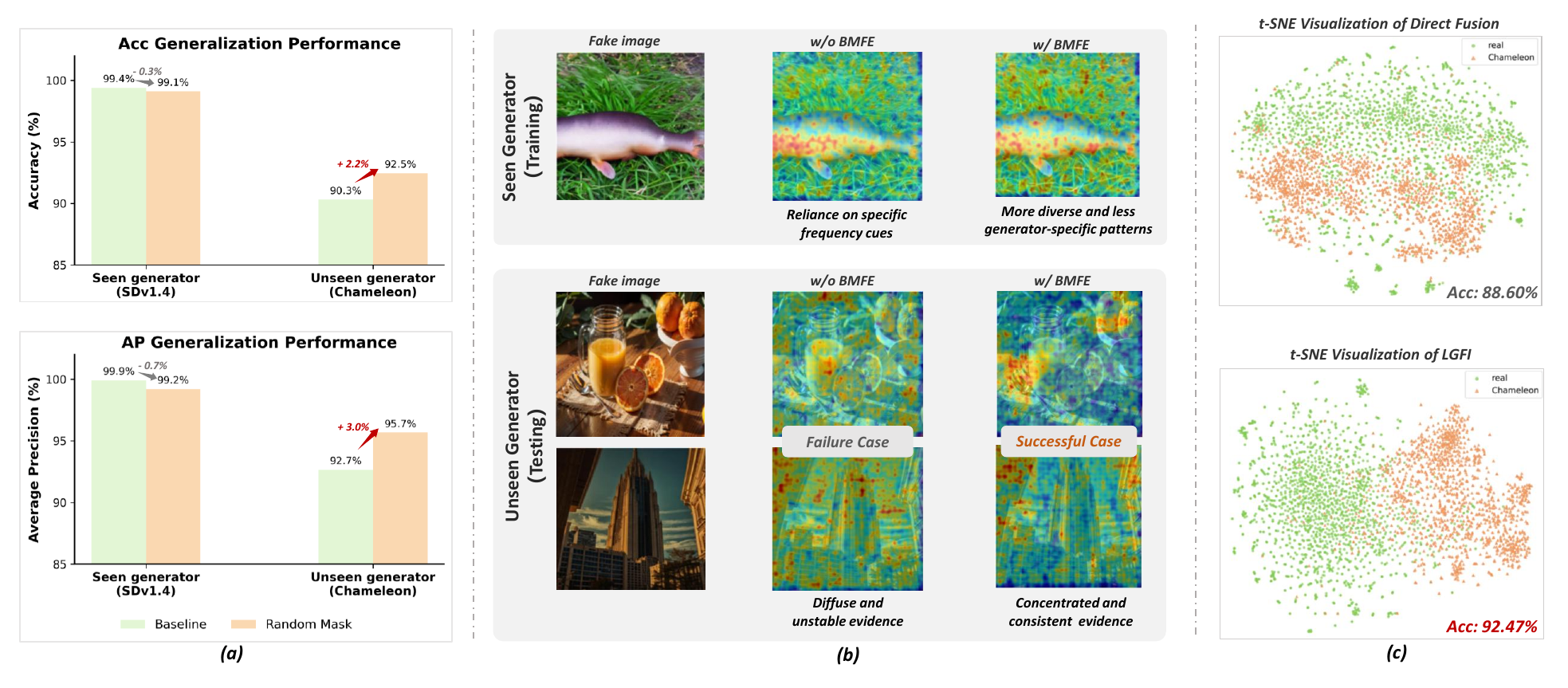}
\vspace{-9mm}
\caption{
 (a) Randomly masking frequency bands narrows the performance gap between seen and unseen generators, revealing a shortcut tendency toward generator-specific frequency cues. (b) Saliency visualization of BMFE. During training (top), models lacking BMFE rely on specific frequency cues, whereas BMFE encourages more diverse patterns. During testing (bottom), such models exhibit diffuse and unstable evidence under distribution shifts, while BMFE produces more concentrated and consistent evidence. (c) t-SNE visualization. Direct fusion produces mixed representations with unclear class separation (top), while LGFI yields more compact and well-separated representations (bottom), improving detection performance.
}

  \label{fig:intro}
\end{figure*}

However, existing methods still exhibit limited cross-generator generalization: they perform well on seen generators but degrade significantly on unseen ones. We attribute this issue to two key limitations. \textit{(1) Frequency Shortcut Bias.} During training, models tend to rely on easily distinguishable frequency cues associated with specific generators, which do not generalize well across sources and lead to degraded performance under distribution shifts. To investigate this, we randomly mask frequency bands during training to perturb frequency cues. As shown in Figure~\ref{fig:intro}(a), masking slightly reduces seen performance but improves unseen performance and narrows the gap, revealing a shortcut tendency toward generator-specific frequency cues. \textit{(2) Cross-Domain Representation Conflict.} Existing dual-stream frameworks typically adopt direct fusion strategies, such as concatenation or cross-attention, to combine frequency cues with semantic representations from VFMs. However, these features reside in inherently different representation spaces: frequency features capture low-level statistical irregularities, while VFM features encode high-level semantic abstractions. This mismatch leads to mixed representations under direct fusion, degrading detection performance.

To address these challenges, we propose a \textbf{F}requency-aware \textbf{G}ated \textbf{I}njection \textbf{N}etwork (\textbf{FGINet}), which aims to learn generalizable frequency representations and integrate them with semantic features in a structure-aware and hierarchical manner. We first design a Band-Masked Frequency Encoder (\textbf{BMFE}), which applies cross-band independent masking to reduce reliance on generator-specific cues and encourage more diverse and generator-invariant frequency representations. As shown in Figure~\ref{fig:intro}(b), during training, the model without BMFE exhibits attention concentrated along specific structures, whereas BMFE leads to a more balanced distribution covering richer frequency patterns. At test time, without BMFE, the model exhibits diffuse and unstable evidence under distribution shifts, while BMFE produces more concentrated and consistent evidence. To further mitigate cross-modal representation conflict, we introduce a Layer-wise Gated Frequency Injection (\textbf{LGFI}) mechanism. Instead of direct fusion, LGFI progressively injects frequency cues into the VFM backbone following its hierarchical feature abstraction, with a learnable gating function dynamically controlling the contribution at each layer. This design preserves the pretrained semantic structure while enabling effective cross-modal interaction. As shown in Figure~\ref{fig:intro}(c), direct fusion leads to entangled representations with unclear class separation (top), whereas LGFI produces more compact and well-separated joint representations (bottom).

Moreover, we introduce a Hyperspherical Compactness Learning (\textbf{HCL}) framework based on the CosFace objective~\cite{wang2018cosface}. By constraining features onto a normalized hypersphere with a cosine margin, HCL enforces intra-class compactness and inter-class separability, encouraging compact and well-separated representations and improving generalization to unseen generative models.

The main contributions of this paper are as follows:
\begin{itemize}
\item We propose a Frequency-aware Gated Injection Network (FGINet) that addresses frequency shortcut bias and semantic-frequency representation conflict via a Band-Masked Frequency Encoder (BMFE) and a Layer-wise Gated Frequency Injection (LGFI) mechanism.
\item We introduce a Hyperspherical Compactness Learning (HCL) with cosine margin constraints to enforce compact and well-separated representations, improving generalization to unseen generators.
\item Extensive cross-domain experiments show that FGINet achieves state-of-the-art performance and superior generalization across multiple challenging datasets.
\end{itemize}

\section{Related Work}
\label{sec:related_work}

\subsection{High-Frequency Artifact-based Detection}
\label{subsec:frequency_based}

Due to the intrinsic mechanisms of generative models, synthesized images often exhibit distinct high-frequency artifacts that deviate from natural image statistics~\cite{frank2020leveraging,tan2024rethinking,durall2020watchup, li2025optimized, qian2020thinking}. Frequency-based works usually exploit this property using shallow convolutional networks, which are particularly effective at capturing local high-frequency patterns in a computationally efficient manner. Methods such as SAFE~\cite{li2025improving}, NPR~\cite{tan2024rethinking}, and MesoNet~\cite{Afchar2018mesonet} adopt lightweight architectures and achieve strong performance across multiple benchmarks. Li et al.~\cite{li2025improving} suggested applying data augmentation to the training of shallow frequency models, which effectively improves generalization. Reconstruction-based approaches, such as DIRE~\cite{wang2023dire}, leverage reconstruction residuals to expose frequency discrepancies, while MRCL~\cite{zhuang25mrcl} extends this idea with multi-reconstruction and contrastive learning for improved representation learning. Moreover, region selection has been explored to better localize informative frequency cues. TextureCrop~\cite{konstantinidou2025texturecrop} prioritizes high-texture regions based on texture density, whereas PatchCraft~\cite{zhong2023patchcraft} and AIDE~\cite{yansanity} exploit the contrast between rich-texture and smooth regions to enhance detection performance.

\subsection{Pre-trained Model-based Detection}
\label{subsec:Pre-trained_Models}

Pre-trained visual models have significantly improved the generalization and robustness of AI-generated image detection by leveraging rich visual priors from large-scale pre-training. Early work such as UnivFD~\cite{ojha2023towards} shows that simple k-NN and linear probing on CLIP features~\cite{radford2021learning} can already achieve strong cross-generator generalization. Subsequent studies further enhance detection from multiple perspectives. Some works extend beyond vision-only architectures by incorporating the CLIP textual branch, e.g., C2P-CLIP~\cite{tan2025c2p}, Fatformer~\cite{liu2024fatformer}, and MiraGe~\cite{shi2025mirage}. Others focus on improving representation robustness, such as introducing a variational information bottleneck (VIB)~\cite{mahabadi2021vibft} into CLIP~\cite{zhang2025towards}, or exploiting shallow-layer features that preserve fine-grained details for forgery detection~\cite{koutlis2024leveraging}. In addition, data-centric approaches, including DRCT~\cite{chen2024drct}, Rajan et al.~\cite{rajan2025aligned}, and DDA~\cite{chen2025dda}, mitigate dataset bias through data alignment, significantly improving cross-domain performance. HiDA-Net~\cite{mu2025hidanet} further enhances local modeling by aggregating region-level features to capture more comprehensive artifact cues.

\subsection{Joint Frequency and Pre-trained Detection}
\label{subsec:pretrained_based}
Integrating frequency-domain information with semantic representations from pre-trained models has emerged as an effective paradigm for enhancing detection performance. Representative works~\cite{yansanity,zhou2025aigi,cheng2025co} demonstrate that combining these complementary cues can significantly improve robustness. However, existing fusion strategies remain limited. AIDE~\cite{yansanity} adopts simple concatenation to combine frequency and semantic features. Building upon simple concatenation, CO-spy~\cite{cheng2025co} introduces dynamically computed scaling factors to enhance the effectiveness of feature fusion. AIGI-Holmes~\cite{zhou2025aigi} independently projects high-frequency information from NPR~\cite{tan2024rethinking} and semantic information from CLIP into the semantic space of Large Language Model (LLM), where the fusion occurs throughout the inference phase. Despite these efforts, most approaches rely on late fusion mechanisms that treat the two modalities as largely independent. Such designs fail to fully exploit the intrinsic correlations between frequency and semantic representations. We argue that frequency information should instead act as a complementary signal that progressively guides semantic representation learning, motivating a more structured and layer-wise integration strategy.

\begin{figure*}[t]
  \centering
 
  {
        \includegraphics[width=1\linewidth]{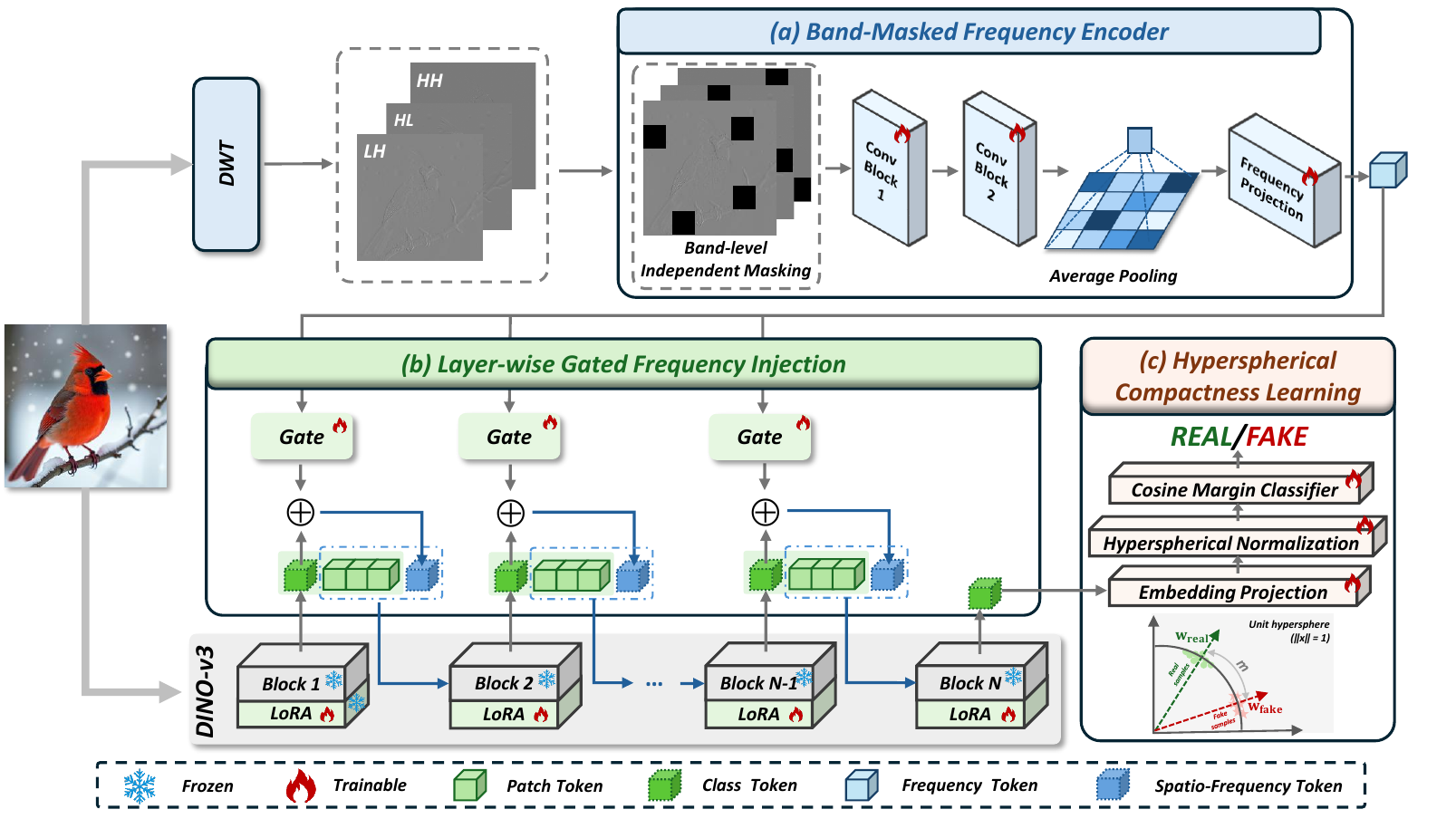}
     }
   \caption{Overview of the proposed Frequency-aware Gated Injection Network (FGINet). FGINet consists of three components: (a) a Band-Masked Frequency Encoder (BMFE) that applies cross-band masking in the frequency domain to learn generalizable frequency representations; (b) a Layer-wise Gated Frequency Injection (LGFI) module that progressively injects frequency cues into a DINOv3 backbone with LoRA, with a learnable gating function dynamically controlling the contribution at each layer; and (c) a Hyperspherical Compactness Learning (HCL) module that enforces compact and well-separated feature distributions via cosine margin optimization.}
       % \vspace{-7pt}
   \label{fig:overview}
\end{figure*}

\section{Method}
\label{sec:Method}
\subsection{Overall Architecture}
To address frequency shortcut bias and cross-domain representation conflict, we propose a \textbf{Frequency-aware Gated Injection Network (FGINet)}, a dual-stream framework for generalizable AI-generated image detection. As shown in Figure~\ref{fig:overview}, our framework processes an input image through two complementary pathways. The semantic pathway utilizes a pretrained DINOv3 with LoRA to extract high-level semantic cues relevant to forgery detection. The frequency pathway employs a \textbf{Band-Masked Frequency Encoder (BMFE)} (Sec.~\ref {sec:Band-Masked Frequency Encoder}) to learn invariant frequency representations by suppressing generator-specific artifacts. To bridge the two domains, we introduce a \textbf{Layer-wise Gated Frequency Injection (LGFI)} mechanism (Sec.~\ref{sec:Layer-wise Gated Frequency Injection}), which progressively injects frequency tokens into the backbone with learnable gates. Finally, a \textbf{Hyperspherical Compactness Learning (HCL)} module (Sec.~\ref{Hyperspherical Compactness Learning}) enforces compact and well-separated representations for both real and fake samples, improving generalization to unseen generators.

\subsection{Band-Masked Frequency Encoder}
\label{sec:Band-Masked Frequency Encoder}
To mitigate frequency shortcut bias, where models tend to over-rely on easily distinguishable generator-specific cues, we introduce a Band-Masked Frequency Encoder (BMFE) to promote more generalizable frequency representations.

\subsubsection{Haar Wavelet Decomposition}
Given an input image $\mathbf{X} \in \mathbb{R}^{C \times H \times W}$, we apply a 2D discrete wavelet transform (DWT) with fixed Haar filters to decompose it into four sub-bands:
\begin{equation}
[\mathbf{X}_{LL}, \mathbf{X}_{LH}, \mathbf{X}_{HL}, \mathbf{X}_{HH}] = \mathrm{DWT}(\mathbf{X}).
\end{equation}

High-frequency artifacts introduced during upsampling and decoding are primarily captured by the $\mathbf{X}_{LH}$, $\mathbf{X}_{HL}$, and $\mathbf{X}_{HH}$ bands, while $\mathbf{X}_{LL}$ encodes low-frequency structure~\cite{frank2020leveraging}. Therefore, we focus on the high-frequency components and discard $\mathbf{X}_{LL}$, which largely contains redundant semantic information already captured by the semantic pathway. The remaining high-frequency components are then concatenated as:
\begin{equation}
\mathbf{X}_{HF} = \mathrm{Concat}(\mathbf{X}_{LH}, \mathbf{X}_{HL}, \mathbf{X}_{HH}) \in \mathbb{R}^{3C \times H_f \times W_f},
\end{equation}
where $H_f = H/2$ and $W_f = W/2$ denote the spatial dimensions of the high-frequency feature maps.

\subsubsection{Band-level Independent Masking} To perturb frequency cues and promote more diverse representations, we apply a Band-level Independent Masking strategy. Given $\mathbf{X}_{HF}$, we partition each frequency band into non-overlapping patches of size $S \times S$, forming a coarse patch grid of size $H_p \times W_p$, where $H_p = H_f / S$ and $W_p = W_f / S$. On this grid, we independently sample a binary mask for each high-frequency band:
\begin{equation}
\mathbf{M} \in \{0,1\}^{3 \times H_p \times W_p}, \quad
M_{b,i,j} \sim \mathrm{Bernoulli}(1 - \rho),
\end{equation}
where $b \in \{1,2,3\}$ indexes the three high-frequency sub-bands, $(i,j)$ denote patch indices, and $\rho \in [0,1]$ is the mask ratio.

Each sampled mask is shared across the RGB channels within the same frequency band. We thus expand $\mathbf{M}$ along the channel dimension to obtain $\hat{\mathbf{M}} \in \{0,1\}^{3C \times H_p \times W_p}$. The mask is then upsampled via nearest-neighbor interpolation to match the spatial resolution of $\mathbf{X}_{HF}$, such that each binary entry corresponds to an $S \times S$ local region. The masked representation is given by:
\begin{equation}
\tilde{\mathbf{X}}_{HF} = \mathbf{X}_{HF} \odot \mathrm{Upsample}(\hat{\mathbf{M}}),
\end{equation}
where $\odot$ denotes element-wise multiplication.

\subsubsection{Frequency Token Encoding}

To inject frequency cues into the Transformer-based semantic pathway, we encode the masked high-frequency map $\tilde{\mathbf{X}}_{HF}$ into a compact frequency token. 

We first employ a lightweight convolutional stem $\Phi_{\text{conv}}$, which is structured with two sequential convolutional blocks (Conv Block 1 and Conv Block 2). Within this stem, each individual block is uniformly composed of a convolutional layer, followed by BatchNorm and ReLU activation:
\begin{equation}
\mathbf{Z} = \Phi_{\text{conv}}(\tilde{\mathbf{X}}_{HF}) \in \mathbb{R}^{C' \times H' \times W'},
\end{equation}
where $\mathbf{Z}$ represents the intermediate frequency representation, and $C'$, $H'$, and $W'$ denote the channel dimension and spatial dimensions after convolution.

We then perform Average Pooling over $\mathbf{Z}$ to obtain a channel-wise representation $\mathbf{v} \in \mathbb{R}^{C'}$. For the $c$-th channel, this aggregation is computed as:
\begin{equation}
v_c = \frac{1}{H'W'} \sum_{i=1}^{H'} \sum_{j=1}^{W'} Z_{c,i,j},
\end{equation}
where $Z_{c,i,j}$ denotes the activation at channel $c$ and spatial location $(i,j)$, and $c \in \{1, \dots, C'\}$ indexes the channel dimension.

Finally, to bridge the dimensional gap between the frequency and semantic streams, channel-wise representation $\mathbf{v}$ is projected into the Transformer embedding space to obtain the frequency token $\mathbf{T}_{freq}$:
\begin{equation}
\mathbf{T}_{freq} = \mathcal{P}(\mathbf{v}),
\end{equation}
where $\mathcal{P}(\cdot)$ denotes the Frequency Projection module, implemented as a linear layer followed by dropout and LayerNorm, to align the feature distribution with the Transformer embedding space.

 \subsection{Layer-wise Gated Frequency Injection}
 \label{sec:Layer-wise Gated Frequency Injection}
To enable effective interaction between the low-level frequency representations and high-level semantic representations, we propose a Layer-wise Gated Frequency Injection (LGFI) mechanism. Instead of employing a direct fusion strategy, LGFI progressively injects frequency cues into the Vision Foundation Model (VFM) backbone, facilitating hierarchical feature abstraction while preserving the pre-trained semantic space.

\subsubsection{Semantic Pathway Adaptation}
The semantic pathway is built upon a pre-trained DINOv3 backbone. To adapt it to the forgery detection task with minimal parameter overhead, we employ Low-Rank Adaptation (LoRA)~\cite{hu2022lora}. Specifically, the backbone weights are kept frozen, and trainable low-rank matrices are injected into the Query, Key, and Value (QKV) projection layers of the Multi-Head Self-Attention (MHSA) blocks. This design enables task-specific adaptation while preserving the generalizable representation of the foundation model, and avoids overfitting by limiting the number of trainable parameters.

\subsubsection{Gated Injection Mechanism}

While the semantic pathway extracts spatial inconsistencies, the frequency token $\mathbf{T}_{freq}$ carries explicit generator-specific artifacts. To bridge these two domains, we dynamically inject $\mathbf{T}_{freq}$ into the class token of the Transformer blocks via learnable layer-wise gated modulation. 

Let $\mathbf{t}_{cls}^{(l)}$ denote the class token at the output of the $l$-th Transformer block. The layer-wise gated injection is formulated as:
\begin{equation}
\hat{\mathbf{t}}_{cls}^{(l)} = \mathbf{t}_{cls}^{(l)} + \alpha^{(l)} \mathbf{T}_{freq},
\end{equation}
where $\alpha^{(l)} \in \mathbb{R}$ is a learnable layer-wise gate that controls the strength of frequency injection. To maintain stability of the pre-trained semantic space during early training, $\alpha^{(l)}$ is initialized to a small constant. The updated token $\hat{\mathbf{t}}_{cls}^{(l)}$ is then passed to the $(l+1)$-th Transformer block together with the patch tokens.

This layer-wise gating adaptively regulates frequency contributions across hierarchical semantic levels. After passing through all $N$ Transformer blocks, we obtain the final spatio-frequency representation $\mathbf{t}_{cls}^{(N)}$, which serves as the unified global representation for classification.

\subsection{Hyperspherical Compactness Learning}
 \label{Hyperspherical Compactness Learning}
To encourage compact and well-separated representations, we introduce a Hyperspherical Compactness Learning (HCL) based on the CosFace objective~\cite{wang2018cosface}. By constraining features onto a unit hypersphere and enforcing a cosine margin, HCL promotes compact intra-class distributions and clear inter-class separation under a stricter geometric constraint.

Given the spatio-frequency representation $\mathbf{t}_{cls}^{(N)}$ from the LGFI module, we project it via the Embedding Projection to obtain $\mathbf{f}$:
\begin{equation}
\mathbf{f} = \mathcal{E}(\mathbf{t}_{cls}^{(N)}),
\end{equation}
where $\mathcal{E}(\cdot)$ denotes the Embedding Projection, implemented as a linear layer followed by LayerNorm.

Next, in the Hyperspherical Normalization step, both features and classifier weights are mapped onto the unit hypersphere by $\ell_2$ normalization:
\begin{equation}
\mathbf{x}_i = \frac{\mathbf{f}_i}{\|\mathbf{f}_i\|_2}, \quad
\mathbf{w}_j = \frac{\mathbf{W}_j}{\|\mathbf{W}_j\|_2},
\end{equation}
where $\mathbf{f}_i \in \mathbb{R}^d$ denotes the feature of the $i$-th sample, $\mathbf{W}_j \in \mathbb{R}^d$ is the weight vector of the $j$-th class (real and fake). The cosine similarity is computed as $\cos \theta_{j,i} = \mathbf{w}_j^\top \mathbf{x}_i$.

Finally, we adopt an additive Cosine Margin Classifier, which enforces a margin between classes on the hypersphere. The loss is formulated as:
\begin{equation}
\mathcal{L}_{HCL} = -\frac{1}{B} \sum_{i=1}^{B} 
\log 
\frac{e^{s(\cos \theta_{y_i,i} - m)}}
{e^{s(\cos \theta_{y_i,i} - m)} + \sum_{j \neq y_i} e^{s \cos \theta_{j,i}}},
\end{equation}
where $B$ is the batch size, $y_i$ denotes the ground-truth class label of the $i$-th sample, $s > 0$ is a scaling factor for logits, and $m > 0$ is the additive cosine margin.

By enforcing feature normalization with an additive cosine margin, HCL organizes features into a compact and well-separated structure on the hypersphere. In practice, real samples form tighter clusters, while features from different classes are separated by a larger margin, leading to improved generalization to unseen generative models.

\begin{table*}[t]
\caption{Performance comparison on the GenImage \citep{zhu2023genimage} dataset using accuracy (Acc) as the metric.}
\vspace{-3mm}
\small \centering
    \renewcommand{\arraystretch}{1}
 \setlength{\tabcolsep}{9pt}
    \scalebox{0.96}[0.96]{
\begin{tabular}{l|cccccccc|c}
\toprule
\textbf{Method}  &{\textbf{Midjourney}} &\textbf{{SD v1.4}} & \textbf{{SD v1.5}} & \textbf{{ADM}} &\textbf{{GLIDE}} &\textbf{{Wukong}} &\textbf{{VQDM} }&\textbf{{BigGAN}} & {\B{\textit{mAcc (\%)}}}\\ \midrule

% \tablecite{CNNDetect}{CVPR 20}{wang2020cnn} & 50.1  & 50.3 & 50.3 & 53.0  & 51.7  & 51.4  & 50.0  & 69.8   & 53.3  \\
% \tablecite{LGrad}{CVPR 23}{tan2023learning}  & 73.7  & 76.3  & 77.4  & 51.8   & 49.8  & 73.1   & 52.1  & 40.5   & 61.8 \\
\tablecite{UniFD}{CVPR 23}{ojha2023towards} & 56.9  & 65.1 & 64.7  & 69.2  & 60.1 & 73.5  & 86.0  & 89.3  & 70.6\\
\tablecite{DIRE}{CVPR 23}{wang2023dire} & 50.4 & 100.0 & 99.9 & 52.5 & 62.7 & 56.5 & 52.4 & 59.5 & 71.2 \\
\tablecite{NPR}{CVPR 24}{tan2024rethinking}  & 77.8  & 78.6 & 78.9  & 69.7   & 78.4  & 76.1  & 78.1  & 80.1   & 77.2 \\
\tablecite{Fatformer}{CVPR 24}{liu2024fatformer} & 92.7 & 100.0 & 99.9 & 75.9 & 88.0 & 99.9 & 98.8 & 55.8 & 88.9 \\
\tablecite{DRCT}{ICML 24}{chen2024drct} & 91.5 & 95.0 & 94.4 & 79.4 & 89.2 & 94.7 & 90.0 & 81.7 & 89.5 \\
% \tablecite{FreqNet}{AAAI 24}{tan2024frequency} & 69.7  & 64.2  & 64.9  & 83.5   & 81.2  & 57.8   & 81.4  & 90.5   & 74.1  \\
% \tablecite{ATTSD}{ICCV 25}{cai2025adaptive} & 91.8 & 94.7 & 93.8 & 85.0 & 97.3 & 91.4 & 81.2 & 90.2 & 90.7 \\
\tablecite{AIDE}{ICLR 25}{yansanity} & 81.4  & 99.8  & 99.8  & 78.5  & 91.8  & 98.9  & 80.2  & 66.8  & 87.2 \\
\tablecite{SAFE}{KDD 25}{li2025improving} & 95.2 & 99.4 & 99.3  & 82.2  & 96.2  & 98.1   & 96.2  & 97.7   & 95.5 \\
\tablecite{VIB}{CVPR 25}{zhang2025towards}  & 88.3   & 99.5  & 99.3  & 74.3  & 73.4  & 98.7   & 89.4  & 55.6  & 84.8  \\
\tablecite{MiraGe}{MM 25}{shi2025mirage} & 83.2 & 98.8 & 98.5 & 82.7 & 91.3 & 97.6 & 92.4 & 96.5 & 92.6 \\
\tablecite{MRCL}{MM 25}{zhuang25mrcl} & 87.2 & 99.6 & 99.6 & 79.8 & 92.2 & 99.6 & 96.5 & 98.7 & 94.2 \\

\rowcolor{lightblue}
\midrule
\textit{\B{FGINet (Ours)}} &92.8  & 99.6  & 99.2  & 89.5   & 98.0 & 99.9   & 99.3 & 95.9  & \textbf{96.7} \\ 
\bottomrule
\end{tabular}
}

\label{table:b1}

\end{table*}

\begin{table*}[t]
\caption{Performance comparison on the Synthbuster~\cite{synthbuster} dataset using accuracy (Acc) as the metric.} 
\vspace{-3mm}
\small \centering
    \renewcommand{\arraystretch}{1}
 \setlength{\tabcolsep}{7pt}
    \scalebox{0.96}[0.96]{
\begin{tabular}{l|ccccccccc|c}
\toprule
\textbf{Method}  &{\B{DALL·E 2}} &\B{{DALL·E 3}} & \B{Firefly} & \B{GLIDE} &\B{Midjourney} &\B{SD 1.3} &\B{SD 1.4} & \B{SD 2} & \B{SDXL} &{\B{\textit{mAcc (\%)}}} \\ \midrule

\tablecite{UniFD}{CVPR 23}{ojha2023towards} & 83.5 & 47.4 & 89.9 & 53.3 & 52.5 & 70.4 & 69.9 & 75.7 & 68.0 & 67.8  \\
\tablecite{NPR}{CVPR 24}{tan2024rethinking} & 51.1 & 49.3 & 46.5 & 48.5 & 52.8 & 51.4 & 51.8 & 46.0 & 52.8 & 50.0 \\
\tablecite{Fatformer}{CVPR 24}{liu2024fatformer} &  59.4 & 39.5 & 60.3 & 72.7 & 44.4 & 53.7 & 54.0 & 52.3 & 69.1 & 56.1\\
\tablecite{DRCT}{ICML 24}{chen2024drct} &  77.2 & 86.6 & 84.1 & 82.6 & 73.7 & 86.6 & 86.6 & 83.2 & 71.3 & 84.8\\
\tablecite{AIDE}{ICLR 25}{yansanity} & 34.9 & 33.7 & 24.8 & 65.0 & 57.5 & 74.1 & 73.7 & 53.2 & 68.4 & 53.9\\
\tablecite{AlignedForensics}{ICLR 25}{rajan2025aligned} & 50.2 & 48.9 & 51.7 & 53.5 & 98.7 & 98.8 & 98.8 & 98.6 & 97.3 & 77.4\\
\tablecite{C2P-CLIP}{AAAI 25}{tan2025c2p} & 55.6 & 63.2 & 59.5 & 86.7 & 52.9 & 75.2 & 76.7 & 69.2 & 77.7 & 68.5  \\
\tablecite{SAFE}{KDD 25}{li2025improving} & 58.0 & 9.9 & 10.3 & 52.2 & 56.7 & 59.4 & 59.1 & 53.0 & 59.5 & 46.5 \\
\tablecite{DDA}{NIPS 25}{chen2025dda} & 86.3 & 90.0 & 91.9 & 76.5 & 93.5 & 92.9 & 92.7 & 93.3 & 93.5 & 90.1 \\
% \tablecite{WaRPAD}{NIPS 25}{choi2025warpad} \\
\rowcolor{lightblue}
\midrule
\textit{\B{FGINet (Ours)}} & 91.9 & 92.3 & 92.9 & 92.3 & 96.1 & 95.1 & 96.4 & 95.7 & 96.0 & \textbf{94.3}  \\
\bottomrule
\end{tabular}
}

\label{table:b2}
% \vspace{-6mm}
\end{table*}

\begin{table}[h]
\caption{Performance comparison on the \textbf{WildRF}~\cite{cavia2024wildRF} dataset using accuracy (Acc) as the metric.}
\vspace{-3mm}
\small \centering
\renewcommand{\arraystretch}{1}
\setlength{\tabcolsep}{4pt}
% The key change is here: remove the '|' before the last 'c'
\scalebox{0.96}[0.96]{
\begin{tabular}{l|ccc|c} 
\toprule
\multirow{2}{*}{\textbf{Method}} & \multicolumn{4}{c}{\textbf{WildRF}} \\ % Correct multicolumn span to 4
\cmidrule(lr){2-5} % Adjust cmidrule to span columns 2 to 5
& \textbf{Facebook} & \textbf{Reddit} & \textbf{Twitter} & {\B{\textit{mAcc (\%)}}} \\ 
\midrule
\tablecite{UniFD}{CVPR 23}{ojha2023towards} & 49.1 & 60.2 & 56.5 & 55.3 \\
\tablecite{NPR}{CVPR 24}{tan2024rethinking} & 78.1 & 61.0 & 51.3 & 63.5 \\
\tablecite{Fatformer}{CVPR 24}{liu2024fatformer} & 54.1 & 68.1 & 54.4 & 58.9 \\
\tablecite{DRCT}{ICML 24}{chen2024drct} & 46.6 & 53.1 & 55.2 & 50.6 \\
\tablecite{AIDE}{ICLR 25}{yansanity} & 57.8 & 71.5 & 45.8 & 58.4 \\
\tablecite{Aligned}{ICLR 25}{rajan2025aligned} & 89.4 & 69.1 & 81.8 & 80.1 \\
\tablecite{C2P-CLIP}{AAAI 25}{tan2025c2p} & 54.4 & 68.4 & 55.9 & 59.6 \\
\tablecite{SAFE}{KDD 25}{li2025improving} & 50.9 & 74.1 & 37.5 & 57.2 \\
\tablecite{DDA}{NIPS 25}{chen2025dda} & 93.1 & 86.4 & 91.5 & 90.3 \\
\rowcolor{lightblue}

\midrule
\textit{\textbf{FGINet (Ours)}} & 95.9 & 90.3 & 95.3 & \B{93.8} \\
\bottomrule

\end{tabular}}
\label{table:wildrf}
\end{table}

\section{Experiments}\label{Experiments}
\subsection{Experimental Setup}\label{Setup}
\noindent  \textbf{Dataset}. Following the training setting of GenImage~\cite{zhu2023genimage}, all our experiments are trained on a dataset constructed from Stable Diffusion v1.4-generated images~\cite{rombach2022sd} and a subset of ImageNet~\cite{russakovsky2015imagenet}. In order to comprehensively evaluate our model's generalization capability and benchmark it against existing approaches, we adopt a series of advanced datasets for testing. These include \textbf{GenImage}~\cite{zhu2023genimage}, a dataset consisting of 8 different generators and an equal number of real images from ImageNet. \textbf{SynthBuster}~\cite{synthbuster} is also utilized, where real samples are sourced from uncompressed images in the Raise dataset~\cite{DangNguyen2015raise}, and the fake samples are generated by 9 diffusion generators. To assess our model's performance in the wild, we employ \textbf{WildRF}~\cite{cavia2024wildRF}, where all images are collected from several social media platforms like Facebook, Reddit and Twitter. We also adopt \textbf{Chameleon}~\cite{yansanity}, a challenging dataset sampled from social media and specifically curated to include samples that are particularly hard for human observers to distinguish. We conduct robustness analysis on the \textbf {RRDataset}~\cite{li2025RR}, which uses social media transmission and re-digitization manner such as scanning, re-photographing for augmentation. \textit{For further details, please refer to the supplementary material.}

\noindent \textbf{Metrics}. Following prior AI-generated image detection works~\cite{ojha2023towards,zhuang25mrcl,li2025improving}, we report accuracy (Acc) as the primary metric, using a fixed decision threshold of 0.5 across all datasets. For ablation studies, we additionally report average precision (AP) to provide a more comprehensive evaluation.

\noindent \textbf{Implementation Details.} We train the model using the AdamW optimizer with a learning rate of $1\times10^{-5}$ and a batch size of 64. During training, images are randomly resized and cropped, with random horizontal flipping applied at a probability of 0.5, while during evaluation, images are resized and center-cropped to $224\times224$. We use a pre-trained DINOv3 ViT-L/14 as the backbone. To enable parameter-efficient adaptation, Low-Rank Adaptation (LoRA) is applied to all attention blocks with a rank of 8 and a scaling factor of 16. The feature dimension of the image encoder is set to $D=1024$. For the frequency module, the masking ratio is set to $\rho=0.4$, and the patch size is $S=8$. The layer-wise gating parameter $\alpha^{(l)}$ is initialized to 0.01. We adopt a CosFace classifier with scaling factor $s=30$ and margin $m=0.35$. The model is trained on four NVIDIA RTX 3090 GPUs for only 3 epochs. 

\subsection{Main Results}
\noindent \textbf{Performance on Standard Datasets.}
Table~\ref{table:b1} and Table~\ref{table:b2} present the performance of our proposed FGINet on the GenImage~\cite{zhu2023genimage} and Synthbuster~\cite{synthbuster} datasets, compared with the baseline and existing state-of-the-art methods. \textbf{GenImage} is a large-scale benchmark comprising over one million images generated by diverse models, including BigGAN~\cite{brock2019lbiggan}, ADM~\cite{kingma2014adam}, Midjourney~\cite{midjourney}, \textit{etc.}
On this dataset, FGINet achieves an accuracy of 96.7\%, outperforming all competing methods. This result demonstrates its strong capability in handling diverse generative mechanisms. \textbf{Synthbuster} focuses on high-quality diffusion model generated images, including DALL·E~\cite{ramesh2022dalle2} and the Stable Diffusion series~\cite{rombach2022sd, podell2023sdxl}, posing a more challenging scenario for detection. 
Despite this, FGINet attains an accuracy of 94.3\%, surpassing the second-best method by a margin of 4.2\%. This demonstrates its effectiveness in detecting highly realistic, diffusion-based forgeries. Overall, the strong performance on these benchmarks validates our design.

\noindent \textbf{Performance on Wild Dataset.} Beyond standard benchmarks, we evaluate our method on real-world datasets. As shown in Table~\ref{table:wildrf} and Table~\ref{table:chameleon}, we assess the in-the-wild detection capability of FGINet on WildRF and Chameleon. For \textbf{WildRF}, Table~\ref{table:wildrf} reports performance across three social media sources (Facebook, Reddit, and Twitter). FGINet achieves the best results on all sources and attains the highest mean accuracy (mAcc), demonstrating strong cross-source generalization. \textbf{Chameleon} is a highly challenging benchmark composed of hard samples from social media that are prone to human misclassification. As shown in Table~\ref{table:chameleon}, we report both real and fake accuracies. Notably, fake samples are significantly harder to detect, and most methods exhibit very low fake accuracy. DDA achieves relatively strong performance on fake samples due to its dual-domain data alignment, which reduces spurious correlations by aligning pixel- and frequency-level statistics between real and synthetic images. Despite this, FGINet achieves the best overall performance, reaching 92.5\% accuracy and surpassing DDA by a margin of 7.7\%. These results highlight that FGINet effectively learns generalizable frequency cues and integrates them in a structured manner, enabling reliable detection under challenging real-world conditions.

\begin{table*}[]
\caption{
Performance comparison on \textbf{Chameleon}~\cite{yansanity} using accuracy (Acc). 
Following prior work~\cite{mu2025hidanet}, most results are directly adopted. 
For DDA~\cite{chen2025dda} and SAFE~\cite{li2025improving}, we report reproduced results since class-wise (real/fake) accuracies are unavailable. 
Both overall and class-wise (real/fake) accuracies are reported.
}
\vspace{-3mm}
\small \centering
\renewcommand{\arraystretch}{1}
\setlength{\tabcolsep}{6pt}
\begin{tabular}{l|cccccccccc} 
\toprule
\multirow{2}{*}{\textbf{Method}} & UnivFD~\cite{ojha2023towards} & DIRE~\cite{wang2023dire} & NPR~\cite{tan2024rethinking} & AIDE~\cite{yansanity} & SAFE~\cite{li2025improving} & DDA~\cite{chen2025dda} & HiDA-Net~\cite{mu2025hidanet} & PPL~\cite{yang2026ppl} & \cellcolor{lightblue}{\textit{\textbf{FGINet}}}  \\ 

& {\color{pubblue}\scriptsize (CVPR 23)} & {\color{pubblue}\scriptsize (ICCV 23)} & {\color{pubblue}\scriptsize (CVPR 24)} & {\color{pubblue}\scriptsize (ICLR 25)} & {\color{pubblue}\scriptsize (KDD 25)} & {\color{pubblue}\scriptsize (NIPS 25)} & {\color{pubblue}\scriptsize (ICLR 26)} & {\color{pubblue}\scriptsize (ICLR 26)} & \cellcolor{lightblue}{\textit{\textbf{(Ours)}}} \\ 

\midrule
\textit{\textbf{Real Acc (\%)}} & 93.5 & 95.7  & \textbf{100.0} & 94.4 & 99.6 & 87.8 & 79.4 & 68.1 & \cellcolor{lightblue} 94.4 \\
\textit{\textbf{Fake Acc (\%)}} & 17.7 & 11.9 & 2.4 & 20.3 & 5.5 & 80.8 & 65.5 & 64.7 & \cellcolor{lightblue}\textbf{89.9} \\
\textit{\textbf{Overall Acc (\%)}} & 55.6 & 59.7  & 58.1 & 62.6 & 59.4 & 84.8 & 73.4 & 66.6 & \cellcolor{lightblue}\textbf{92.5}  \\

% 使用 specialrule 替代 bottomrule 和 hhline
\specialrule{\heavyrulewidth}{0pt}{0pt} 
% \vspace{-3mm}
\end{tabular}
\label{table:chameleon}
\end{table*}

\begin{table}[h]
\centering

\caption{Ablation on model components. We evaluate the contributions of BMFE (Band-Masked Frequency Encoder), LGFI (Layer-wise Gated Frequency Injection), and HCL (Hyperspherical Compactness Learning) on the Chameleon dataset~\cite{yansanity}. The baseline denotes LoRA fine-tuning~\cite{hu2022lora} on DINOv3~\cite{siméoni2025dinov3}.}
\label{table:ablation2}

\resizebox{0.92\linewidth}{!}{
\begin{tabular}{l c c c c c}
\toprule
\textbf{Method} & \textbf{BMFE} & \textbf{LGFI} & \textbf{HCL} & \textit{\textbf{Acc (\%)}} & \textit{\textbf{AP (\%)}} \\
\midrule
Baseline  &           &           &           & 81.8 & 91.1 \\
+ BMFE                   & \cmark &           &           & 83.1 & 94.0 \\
+ LGFI                   &           & \cmark &           & 88.8 & 91.3 \\
+ BMFE + LGFI            & \cmark & \cmark &           & 90.6 & 94.5 \\
\rowcolor{lightblue}
+ BMFE + LGFI + HCL & \cmark & \cmark & \cmark & \textbf{92.5} & \textbf{95.7} \\
\bottomrule
\end{tabular}
}

\end{table}

\begin{table}[h]
\centering
\setlength{\tabcolsep}{10pt} % 调整列间距
\renewcommand{\arraystretch}{1} % 调整行高
\caption{Impact of fusion strategies and injection depth. LFI denotes layer-wise frequency injection without gating, while LGFI denotes layer-wise gated frequency injection. All models are trained on the SDv1.4 training set~\cite{zhu2023genimage} and evaluated on the Chameleon dataset~\cite{yansanity}.}
\label{table:ablation}
\resizebox{0.9\linewidth}{!}{
\begin{tabular}{lcccc}
\toprule
\textbf{Method} & \textbf{Layers} & \textbf{Gate} & \textit{\textbf{Acc  (\%)}}& \textit{\textbf{AP  (\%)}} \\
\midrule
\multicolumn{5}{c}{\small\textit{Late Fusion}} \\
\midrule
Add        & last    & \xmark & 89.3 & 92.7 \\
Concat     & last    & \xmark & 88.6 & 91.8 \\
Cross-Attention & last    & \xmark & 89.3 & 93.2 \\
\midrule
\multicolumn{5}{c}{\small\textit{LGFI (ours)}} \\
\midrule
LFI        & all     & \xmark & 68.9 & 69.9 \\
LGFI (early)      & 1-8    & \cmark & 88.6 & 90.4 \\
LGFI (middle)      & 8-16   & \cmark & 82.5 & 86.1 \\
LGFI (late)         & 16-24  & \cmark & 83.1 & 88.1 \\
\midrule
\rowcolor{lightblue}
\textbf{Full LGFI} & \textbf{all} & \cmark & \textbf{92.5} & \textbf{95.7} \\
\bottomrule
 \vspace{-6mm}
\end{tabular}}
\end{table}

\subsection{Ablation Studies}

\noindent \textbf{Ablation on Model Components.} We conduct a component-wise ablation study on the Chameleon dataset to evaluate the effectiveness of each module in FGINet, including the Band-Masked Frequency Encoder (BMFE), Layer-wise Gated Frequency Injection (LGFI), and Hyperspherical Compactness Learning (HCL). As shown in Table~\ref{table:ablation2}, the baseline (LoRA fine-tuned DINOv3) achieves 81.8\% Acc and 91.1\% AP. BMFE consistently improves performance, suggesting that perturbing frequency cues reduces reliance on generator-specific patterns. LGFI brings larger gains (over 7\% Acc), highlighting the effectiveness of progressive gated injection over direct fusion. Combining BMFE and LGFI further boosts performance, indicating their complementary roles. Finally, incorporating HCL achieves the best results (92.5\% Acc and 95.7\% AP), showing that hyperspherical compactness enhances feature discriminability. Overall, all components contribute consistently and yield significant cumulative improvements.

\noindent \textbf{Analysis of Injection Strategy and Depth.}
We provide a finer-grained analysis of LGFI by examining different injection strategies and layer-wise injection depths. All other components are kept fixed, and experiments are conducted on the Chameleon dataset. As shown in Table~\ref{table:ablation}, direct fusion methods (Add, Concat and Cross-Attention) applied at the final layer yield limited gains, indicating ineffective alignment between frequency and semantic representations. layer-wise frequency injection (LFI) without gating leads to clear performance degradation, suggesting that indiscriminate injection disrupts the pretrained semantic structure. In contrast, LGFI consistently outperforms both direct fusion and ungated injection. Among partial strategies, early-layer injection performs best, highlighting the importance of low-level alignment. Applying LGFI across all layers further achieves the best results, demonstrating the advantage of progressive and gated integration throughout the hierarchy.

% \noindent \textbf{Analysis of Frequency Band Selection.}
% We analyze the impact of different frequency band selections on the Chameleon dataset. As shown in Table~\ref{tab:ablation_bands}, using only low-frequency (LL) or high-frequency (HH) components yields inferior performance, indicating that a single band provides limited cues. Notably, removing the low-frequency component (w/o LL) achieves the best performance. This suggests that forgery artifacts are predominantly reflected in high-frequency components~\cite{frank2020leveraging}, while low-frequency bands mainly encode semantic content and may introduce redundant information. In contrast, using all bands slightly degrades performance, implying that indiscriminate inclusion of frequency bands may dilute informative high-frequency cues. Overall, these results highlight the importance of selectively leveraging frequency bands for robust forgery detection.

% \begin{table}[t]
% \centering
% \caption{Ablation study of different frequency band selections on the \textbf{Chameleon}~\cite{yansanity} dataset.}
% \label{tab:ablation_bands}
% \small
% \setlength{\tabcolsep}{6pt} % 调整列间距
% \renewcommand{\arraystretch}{1.2} % 调整行高
% \resizebox{0.65\linewidth}{!}{
% \begin{tabular}{c|cccc} % 保留左侧一条竖线
% \toprule
% % \rowcolor{lightblue}
% \textbf{Band} & \textbf{LL Only} & \textbf{HH Only} & \textbf{w/o LL} & \textbf{All} \\
% \midrule
% \textit{\textbf{Acc (\%)}} & 90.3 & 90.8 & \textbf{92.5} & 92.0 \\
% \textit{\textbf{AP (\%)}} & 93.0 & 93.5 & \textbf{95.7} & 95.3 \\
% \bottomrule

% \end{tabular}}
% % \vspace{-3mm}
% \end{table}

\begin{figure}[t]
  \centering
 
  {
        \includegraphics[width=0.8\linewidth]{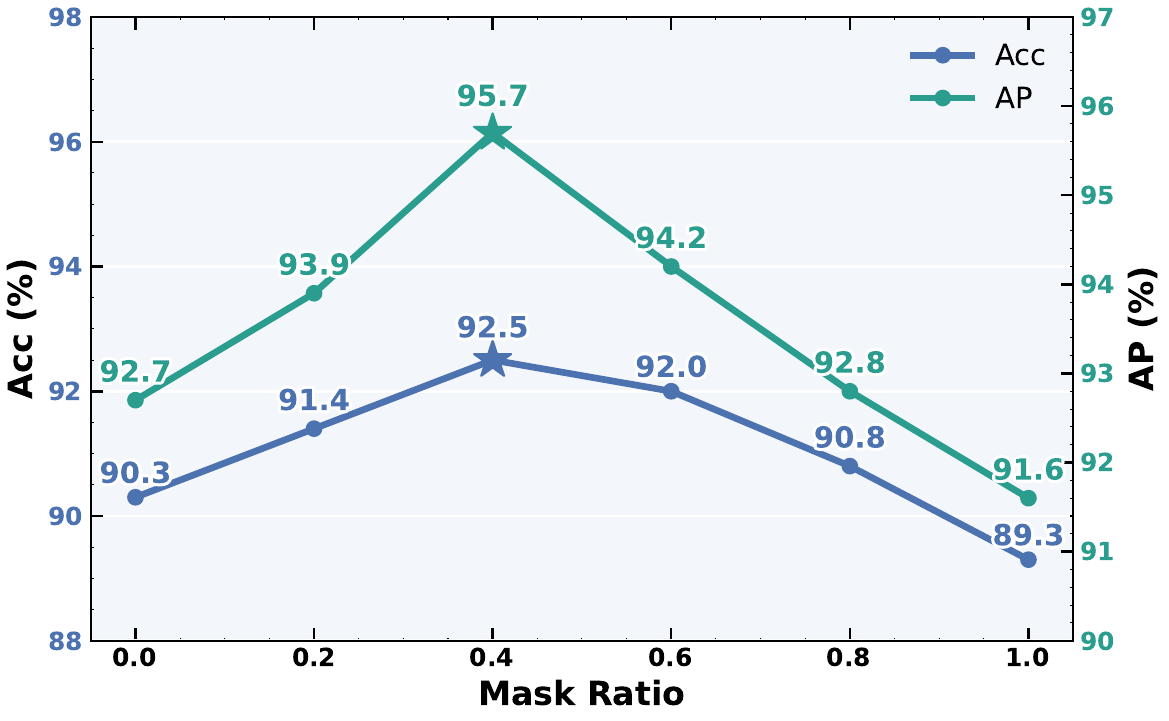}
     }
      \vspace{-3pt}
  \caption{Ablation study on the mask ratio in BMFE. We evaluate the effect of different mask ratios and report both accuracy (Acc) and average precision (AP). Models are trained on SDv1.4~\cite{zhu2023genimage} and evaluated on the Chameleon dataset~\cite{yansanity}.}
       \vspace{-12pt}
   \label{fig:abl_maskratio}
\end{figure}

\noindent \textbf{Impact of Mask Ratio.}
We analyze the effect of different mask ratios in BMFE on the Chameleon dataset. As shown in Figure~\ref{fig:abl_maskratio}, moderate masking yields the best performance, with both Acc and AP peaking at a ratio of 0.4. When the mask ratio is too low, the model still relies on easily distinguishable frequency cues, leading to limited generalization. In contrast, excessive masking degrades performance, as overly aggressive perturbation removes informative frequency structures. These results indicate that an appropriate level of frequency masking effectively balances suppressing shortcut cues and preserving discriminative information, improving generalization. \textit{Please refer to the supplementary material for more details on frequency band selections and LoRA ablation studies.}

\subsection{Visualization Result}
\begin{figure}[t]
  \centering
 
  {
        \includegraphics[width=0.85\linewidth]{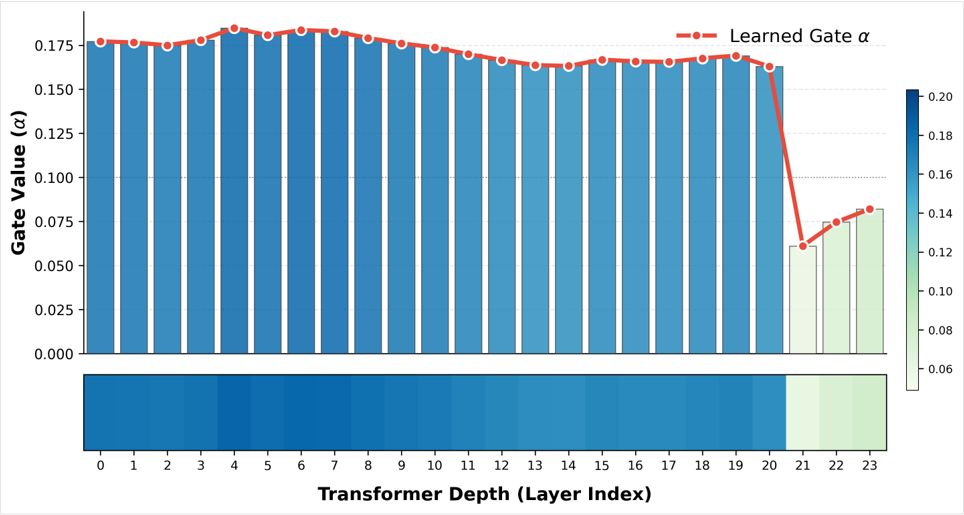}
     }
      \vspace{-3pt}
   \caption{Learned gate visualization. Layer-wise gated frequency injection on DINOv3 enables selective and adaptive control of frequency information across network depth. 
}
      
   \label{fig:gate}
\end{figure}

\noindent \textbf{Learned Gate Visualization.}
We visualize the learned gating values across the Transformer layers of DINOv3 in Figure ~\ref{fig:gate}. The gate values remain relatively high in shallow and intermediate layers, while significantly decrease in deeper layers. This indicates that the model tends to emphasize frequency information at early stages, where low-level structures are more prominent, and progressively suppress it at deeper layers to avoid interfering with high-level semantic representations. Such a pattern is well aligned with our LGFI design, demonstrating that layer-wise gated injection enables adaptive and structure-aware integration of frequency cues across network depth.

\begin{figure}[t]
  \centering
 
  {
        \includegraphics[width=1\linewidth]{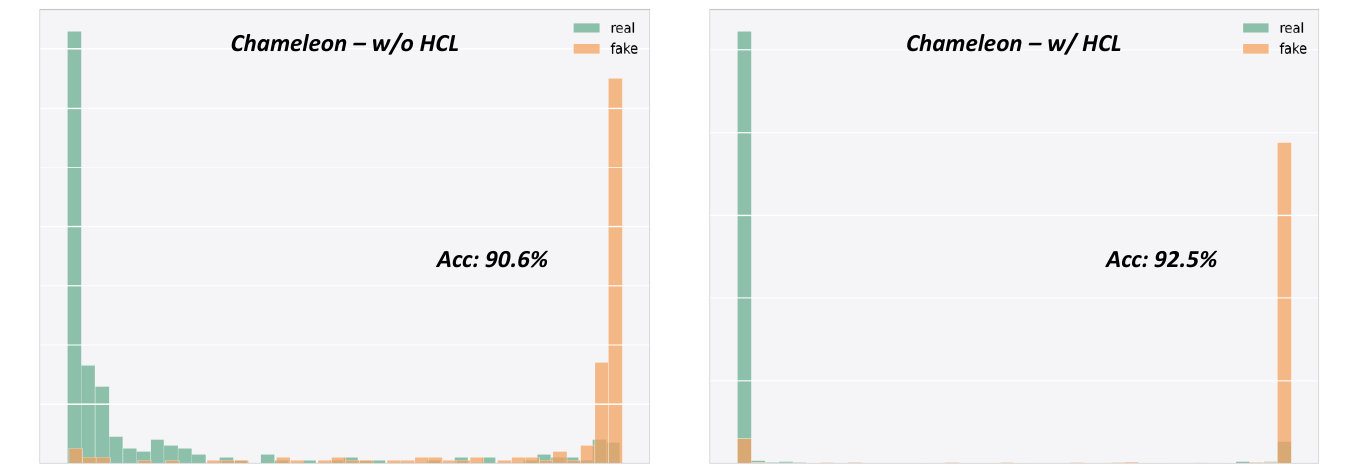}
     }
     \vspace{-8pt}
   \caption{Logit distributions on Chameleon dataset without and with HCL. HCL enforces more compact and well-separated representations, leading to clearer class separation and improved accuracy.}
       \vspace{-5pt}
   \label{fig:tsne}
\end{figure}

\noindent \textbf{Logit Distribution Visualization with HCL.}
We visualize the logit distributions on the Chameleon dataset without and with Hyperspherical Compactness Learning (HCL). Without HCL, the distributions are more dispersed, with overlap in the mid-confidence region, leading to less certain predictions. With HCL, the distributions become more concentrated and better separated, resulting in clearer decision boundaries. 
This improvement also leads to higher detection accuracy, as reflected by the gain from 90.6\% to 92.5\%.

\noindent \textbf{Feature Space Visualization.}
We visualize the feature distributions of our method and the state-of-the-art SAFE on the WildRF dataset, which contains fake images from diverse social media sources (Facebook, Reddit, and Twitter). 
As shown in Figure ~\ref{fig:tsne}, SAFE exhibits overlapping representations, where real samples are heavily mixed with fake samples from different sources. In contrast, our method produces more compact intra-class clusters and clearer separation between real and fake samples, while maintaining better consistency across different domains. 
This indicates that FGINet learns more generalizable representations under cross-source distribution shifts.

\subsection{Robustness Evaluation}

\begin{figure}[t]
  \centering
 
  {
        \includegraphics[width=0.9\linewidth]{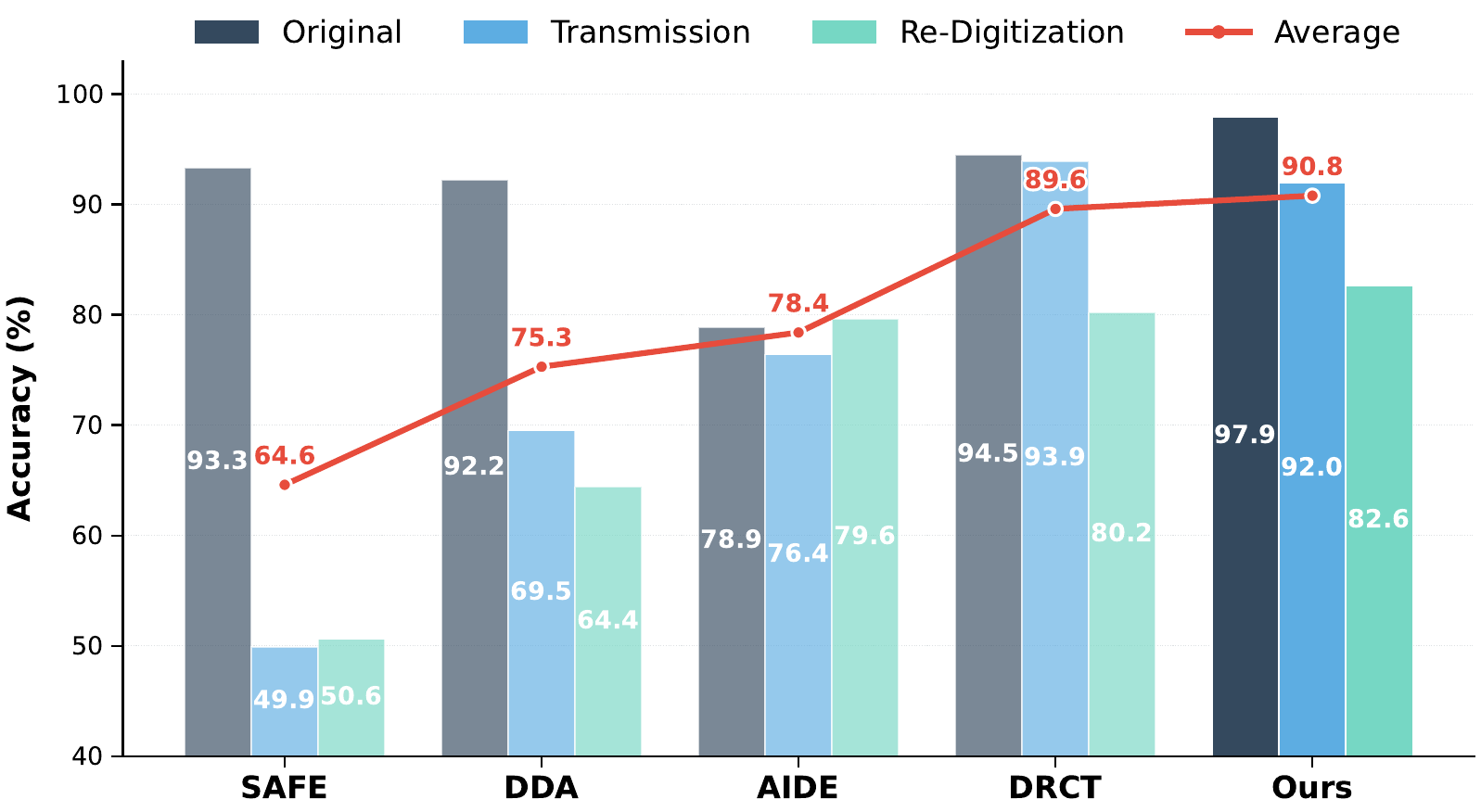}
     }
     \vspace{-5pt}
  \caption{Robustness evaluation on RRDataset~\cite{li2025RR}. Compared with SOTA detectors, our method achieves consistently superior performance under challenging post-processing scenarios, including \textbf{Transmission} and \textbf{Re-Digitization}. The red curve denotes the average accuracy across conditions.}
       \vspace{-7pt}
   \label{fig:rrdataset}
\end{figure}

We further evaluate our method on RRDataset to assess robustness under realistic degradations. Compared to conventional perturbations such as Gaussian blur, compression, or additive noise, RRDataset simulates two common yet more realistic processes: \textbf{Transmission}, involving repeated sharing and compression during internet propagation, and \textbf{Re-Digitization}, caused by scanning or re-photographing. These processes introduce severe quality degradation and distribution shifts, posing significant challenges to reliable detection. As shown in Figure~\ref{fig:rrdataset}, all methods experience notable performance drops under these conditions, highlighting the difficulty of such perturbations. In contrast, our method consistently achieves superior performance across all conditions and attains the highest average accuracy. We attribute this robustness to two key factors. First, LGFI preserves the strong generalization capability of the pretrained DINOv3 by avoiding destructive fusion and enabling adaptive integration of frequency cues. Second, BMFE introduces cross-band random masking, encouraging the model to learn more diverse and generalizable frequency representations rather than relying on specific artifacts. These results demonstrate that FGINet maintains reliable detection performance even under severe real-world degradations.

\section{Conclusion}
In this paper, we study the problem of limited cross-generator generalization in AI-generated image detection. We identify two key limitations: a frequency shortcut bias toward easily distinguishable generator-specific cues, and a representation conflict between low-level frequency and high-level semantic features under direct fusion. To address these issues, we propose FGINet, a frequency-aware framework that learns more generalizable frequency representations and integrates them with semantic features in a structured and hierarchical manner. Specifically, BMFE reduces reliance on generator-specific cues via cross-band masking in the frequency domain, while LGFI enables progressive and adaptive frequency injection aligned with the hierarchical abstraction of Vision Foundation Models (VFMs). HCL further enhances feature compactness and separability through hyperspherical constraints with a cosine margin. Extensive experiments demonstrate that FGINet achieves state-of-the-art performance and strong generalization across multiple challenging datasets.

\begin{figure}[t]
  \centering
 
  {
        \includegraphics[width=0.95\linewidth]{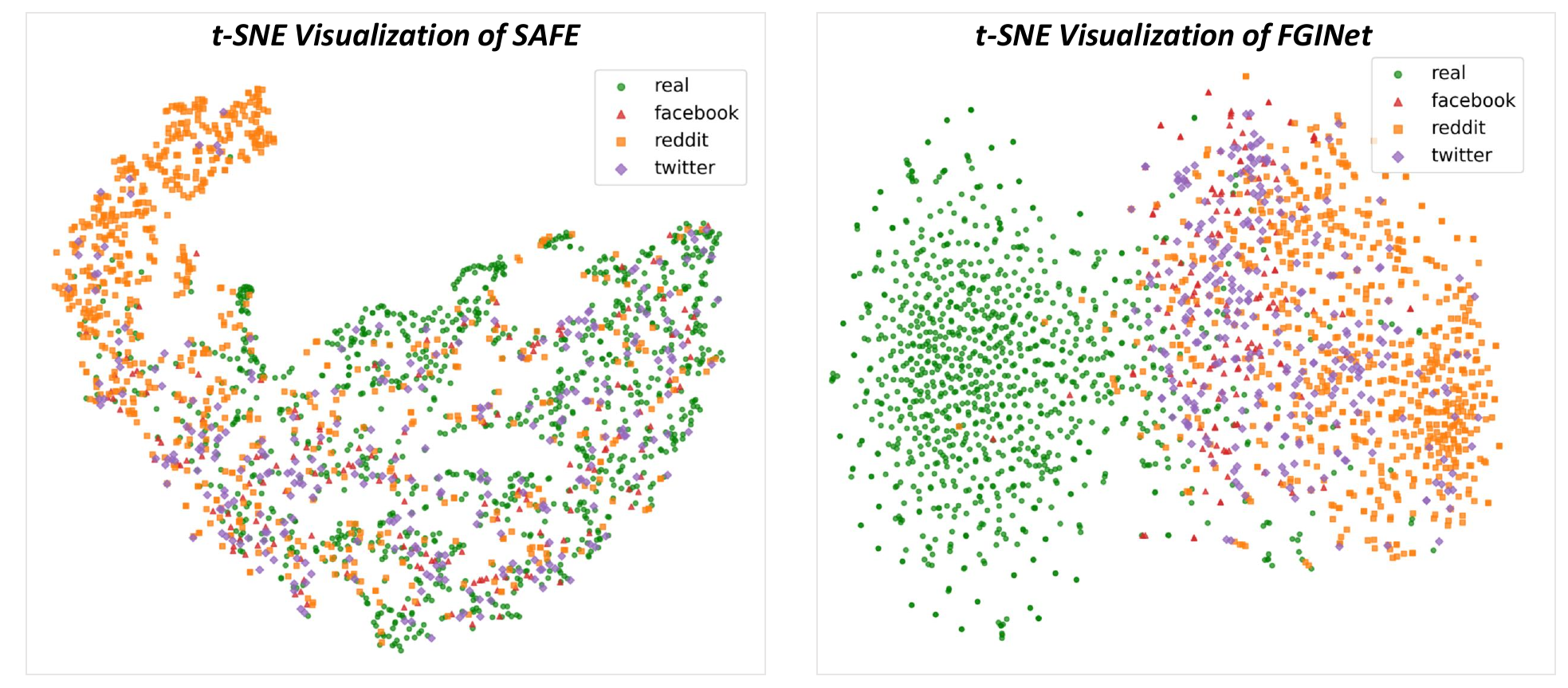}
     }
     \vspace{-2pt}
   \caption{t-SNE visualization comparing SAFE and FGINet on WildRF dataset. FGINet shows more clearer separation between real and fake samples.}
       \vspace{-3pt}
   \label{fig:tsne}
\end{figure}

\begin{acks}
To Robert, for the bagels and explaining CMYK and color spaces.
\end{acks}

%%
%% The next two lines define the bibliography style to be used, and
%% the bibliography file.
\bibliographystyle{ACM-Reference-Format}
\bibliography{sample-base}

%%
%% If your work has an appendix, this is the place to put it.
% \appendix

% \section{Research Methods}

% \subsection{Part One}

% Lorem ipsum dolor sit amet, consectetur adipiscing elit. Morbi
% malesuada, quam in pulvinar varius, metus nunc fermentum urna, id
% sollicitudin purus odio sit amet enim. Aliquam ullamcorper eu ipsum
% vel mollis. Curabitur quis dictum nisl. Phasellus vel semper risus, et
% lacinia dolor. Integer ultricies commodo sem nec semper.

% \subsection{Part Two}

% Etiam commodo feugiat nisl pulvinar pellentesque. Etiam auctor sodales
% ligula, non varius nibh pulvinar semper. Suspendisse nec lectus non
% ipsum convallis congue hendrerit vitae sapien. Donec at laoreet
% eros. Vivamus non purus placerat, scelerisque diam eu, cursus
% ante. Etiam aliquam tortor auctor efficitur mattis.

% \section{Online Resources}

% Nam id fermentum dui. Suspendisse sagittis tortor a nulla mollis, in
% pulvinar ex pretium. Sed interdum orci quis metus euismod, et sagittis
% enim maximus. Vestibulum gravida massa ut felis suscipit
% congue. Quisque mattis elit a risus ultrices commodo venenatis eget
% dui. Etiam sagittis eleifend elementum.

% Nam interdum magna at lectus dignissim, ac dignissim lorem
% rhoncus. Maecenas eu arcu ac neque placerat aliquam. Nunc pulvinar
% massa et mattis lacinia.

\end{document}